\documentclass[letterpaper, 10pt, conference]{ieeeconf}
\IEEEoverridecommandlockouts
\usepackage{cite}
\usepackage{amsmath,amssymb,amsfonts}
\usepackage{algorithmic}
\usepackage{tabularx}
\usepackage{graphicx}
\usepackage{textcomp}
\usepackage{xcolor}
\usepackage{url} 
\usepackage{multirow}
\let\labelindent\relax      
\usepackage{enumitem}       
\usepackage{booktabs}    
\usepackage{caption}     
\usepackage{threeparttable}
\begin{document}


\title{OpenLKA: An Open Dataset of Lane Keeping Assist from Recent Production Vehicles under Real-World Driving Conditions\\}

\author{%
    Yuhang Wang, Abdulaziz Alhuraish, Shengming Yuan, and Hao Zhou\textsuperscript{*}%
    \thanks{%
    All authors are with the Department of Civil and Environmental Engineering, 
    University of South Florida, Tampa, FL 33620, USA
    (e-mail: \{yuhangw, aalhuraish, shengming, haozhou1\}@usf.edu).
    \newline\indent
    \textit{*\,Corresponding author: Hao Zhou. haozhou1@usf.edu}}%
}

\maketitle


\begin{abstract}
Lane Keeping Assist (LKA) is widely adopted in modern production vehicles, yet its real-world performance remains opaque due to the system's proprietary control stack, preventing researchers from diagnosing or improving this technology. To fill the gap, this paper presents OpenLKA, the first open, large-scale dataset for LKA evaluation and improvement. It includes 389.1 hours of LKA-steered data from 62 production vehicle models, collected through extensive road testing in Tampa, Florida and global contributors from the open-source community. The dataset spans a wide range of challenging scenarios, including degraded lane markings, complex road geometries, adverse weather, lighting conditions and various surrounding traffic. The dataset is multimodal, comprising: i) decoded vehicle internal messages including key LKA signals (e.g., system disengagements, lane detection failures); ii) synchronized high-resolution videos from a mounted dash camera; and iii) rich scene annotations generated by a vision language model, covering lane visibility, pavement quality, weather, lighting, and traffic conditions etc. Collectively, OpenLKA provides a comprehensive platform for benchmarking the real-world performance of production LKA systems, identifying safety-critical operational scenarios, and assessing the readiness of current road infrastructure for autonomous driving. The dataset is publicly available at: \url{https://github.com/OpenLKA}.
\end{abstract}

\begin{figure*}[t]
  \centering
  \includegraphics[width=0.78\textwidth]{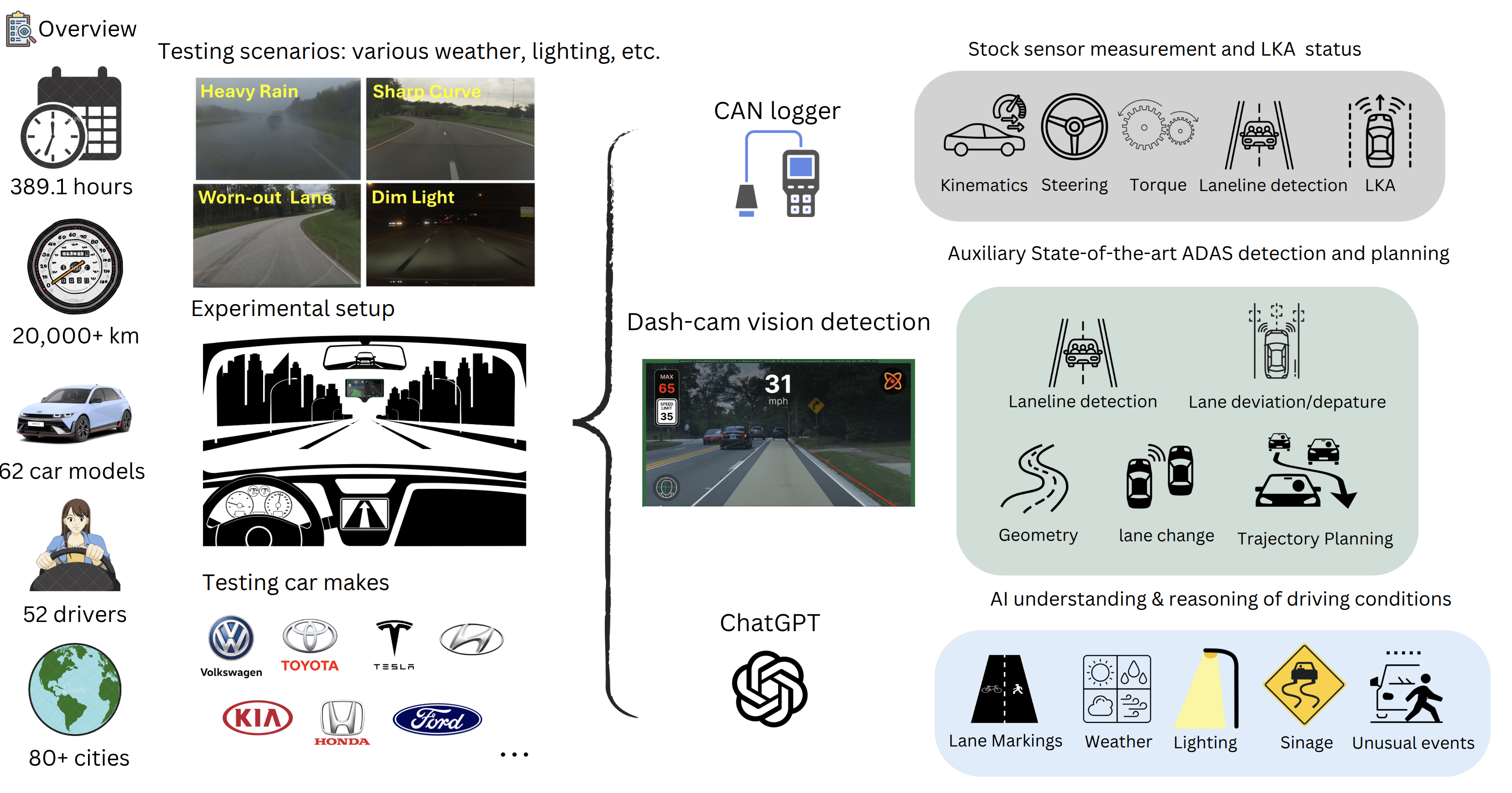}
  \caption{OpenLKA dataset overview\protect\footnotemark. 
}
  \label{fig:lka_overview}
  \vspace{-0.5em}
\end{figure*}

\footnotetext{Logos © respective manufacturers (media-kit downloads, May 2025);
icons © The Noun Project, CC BY 3.0.}

\section{Introduction}
Road-departure crashes account for a substantial share of traffic fatalities, and inadvertent lane-keeping errors are recognised as their primary cause\,\cite{becker2020safety}. To mitigate this risk, automakers have accelerated the roll-out of Advanced Driver Assistance Systems (ADAS)\,\cite{nidamanuri2021progressive,yurtsever2020survey, wienke2012openacc}. Lane-Keeping Assist (LKA) has progressed from a premium option to a default feature, offering continuous steering support intended to keep vehicles centred in their lanes\,\cite{lazcano2021mpc,monfort2022speeding,dean2022estimating}. Yet field reports show that production LKA stacks can drift or disengage on sharp curves, worn lane markings, or in adverse weather\,\cite{waghchoure2022comprehensive}; drivers therefore remain uncertain about when the system will intervene or relinquish control\,\cite{rizgary2024driver}. At the same time, infrastructure agencies lack quantitative guidance on lane-marking contrast or geometric design that accommodates machine vision\,\cite{gopalakrishna2021impacts}, and researchers are hampered by limited access to real-world failure logs—information manufacturers seldom share\,\cite{orieno2024future}.

While a few transportation departments have begun examining LKA behaviour in specific settings\,\cite{pike2024assessing}, such studies involve only a handful of vehicles and roadway conditions. The research community therefore still lacks a large-scale, multimodal corpus centred on LKA—one that spans multiple makes, climates, and driving styles—to support systematic benchmarking and infrastructure evaluation.

\begin{figure}
  \centering
  \includegraphics[width=0.48\textwidth]{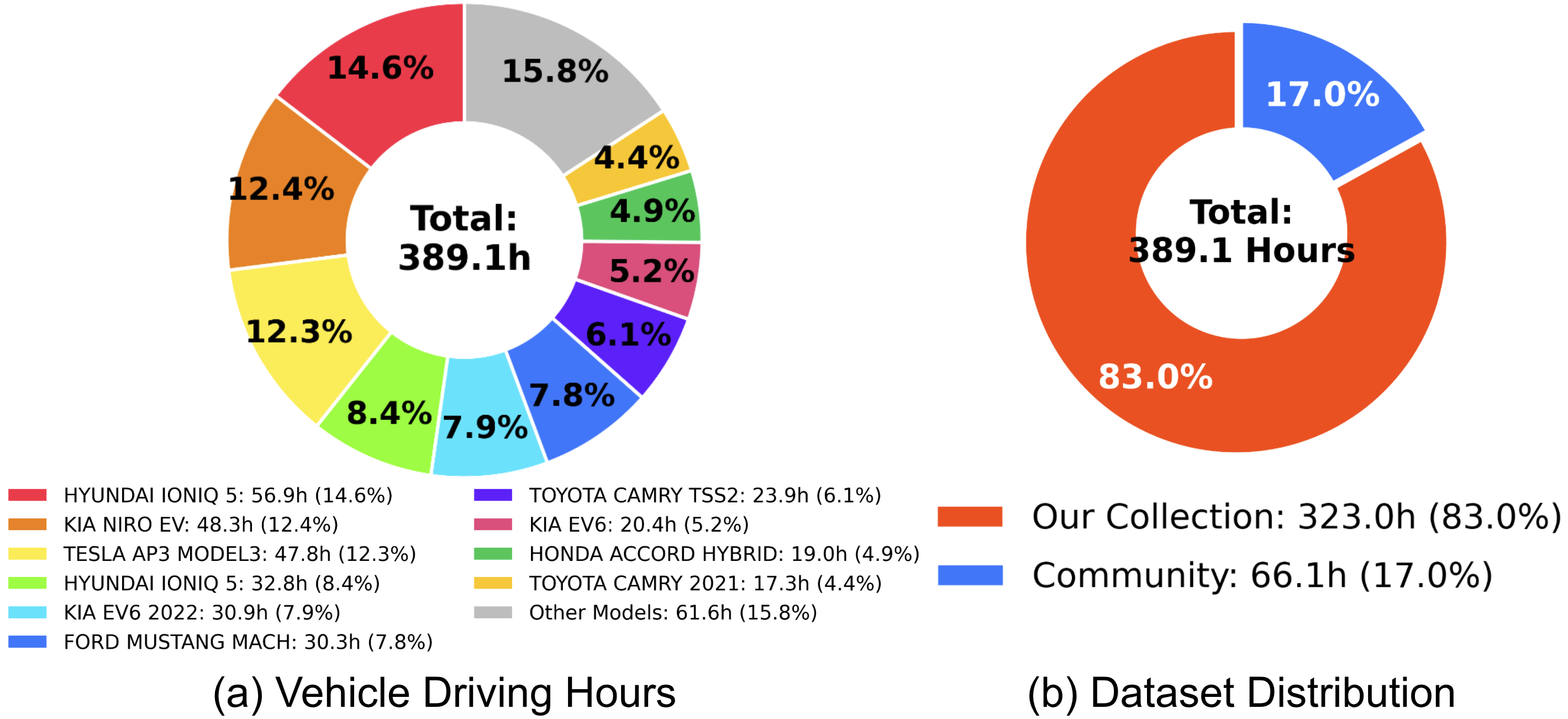}
  \caption{Distribution of hours by car models in OpenLKA}
  \label{fig:DatasetPie}
  \vspace{-0.5em}
\end{figure}

To bridge this gap we introduce \textbf{OpenLKA}; see Fig.\ref{fig:lka_overview};, an openly available dataset that fuses full Controller Area Network (CAN) messages, high-resolution dash-cam video and lane-detection outputs from production vehicles; the corpus combines a controlled campaign—19 vehicle models driven by three professional drivers for 2,986 LKA-engaged trips (19,380 min) in Tampa, Florida—with 775 community-contributed drives (3,964 min) from 52 models that interleave LKA operation and purely human control, thereby yielding in total 62 car models, 52 drivers, 3,761 drives and 389.1 hours of synchronised CAN–video data, a scale that not only exposes brand- and environment-dependent LKA behaviour but also enables direct human-versus-automation comparisons and rich analyses of driver–LKA interaction in everyday traffic, as summarised in Figure~\ref{fig:DatasetPie}.

The contributions of this work are as follows:

\begin{enumerate}[label=(\roman*)]
    \item \textbf{A first-of-its-kind open dataset for real-world LKA analysis:} OpenLKA provides 389.1 hours of multimodal driving data from 62 production vehicle models across various car makes, collected under diverse and challenging real-world conditions, including adverse weather, complex road geometries, and degraded lane markings.
    

    \item \textbf{A unified LKA data collection pipeline and ecosystem:} We introduce a scalable data collection framework that captures full CAN bus streams and synchronized dash-cam video, together with real-time lane detection results from a parallel vision model from Openpilot \cite{openpilot2024}, which enables scalable estimation of road curvature and vehicle in-lane position that would otherwise require costly LIDAR or manual surveys. This pipeline further powers an \textbf{ecosystem} that continuously ingests new LKA and human steering data from global community drivers operating in diverse locations and road conditions.

    \item \textbf{VLM-based rich contextual annotation:} Each driving segment is automatically labeled using Vision Language Models (VLM), with proper prompt engineering, the dataset captures accurate and detailed environmental factors including lane visibility, pavement quality, weather, lighting, and traffic. 
\end{enumerate}


The remainder of the paper is organised as follows. Section II surveys related datasets and LKA field studies. Section III describes the hardware and software pipeline used to collect and synchronise CAN and video data. Section IV presents dataset structure and statistics. Section V explains the annotation and quality-assurance procedures. Section VI outlines research directions that OpenLKA enables, and Section VII concludes with future work.

\section{Related Work}
\label{sec:related_work}

LKA systems, a cornerstone of ADAS, utilize computer vision (CV) techniques to detect lane lines and employ control algorithms to maintain vehicles within lane boundaries, significantly enhancing driving safety \cite{wang2022end}. These systems typically rely on convolutional neural networks (CNNs) to process camera inputs, achieving lane detection accuracies exceeding 97\% on benchmark datasets like TuSimple \cite{tusimple2017}. Despite advancements in CV, lane detection remains vulnerable to long-tail scenarios such as adverse weather conditions, poor lighting, or unconventional lane markings, issues exacerbated by imbalanced training datasets that do not represent diverse global conditions \cite{sultana2023vision}. Moreover, LKA disengagements are not solely attributable to lane detection failures; system designs encompass multiple factors, including sensor fusion and driver interaction, underscoring the necessity of empirical evaluations of market-ready products \cite{zakaria2023lane}. Perception failures can propagate to downstream planning modules, where suboptimal design may impair effective risk mitigation, and control systems occasionally fail to maintain consistent lane centering, posing safety hazards \cite{chalmers2025challenges}. Given the proprietary, black-box nature of LKA systems, their evaluation is particularly crucial through methods such as AI explainability, which can elucidate perception failures and guide infrastructure improvements (e.g., optimized lane markings) to enhance safety and mitigate lane departure accidents—a leading cause of road fatalities \cite{gan2024large}. This synergy between technological capabilities and infrastructure quality highlights the ongoing need for robust, transparent, and comprehensive LKA system research.

Currently, the design and evaluation of ADAS, LKA, and autonomous driving technologies are critically hindered by the lack of comprehensive, multimodal datasets that capture diverse real-world conditions \cite{yurtsever2020survey, nidamanuri2021progressive}. Existing datasets, such as KITTI \cite{geiger2012kitti} and BDD100K \cite{yu2020bdd100k}, primarily focus on sensor and video data but lack crucial CAN bus integration, limiting their applicability in analyzing detailed vehicle dynamics and internal system states critical to understanding LKA behavior. Moreover, many of these datasets are narrowly scoped toward fully autonomous driving scenarios in single specific testing vehicle, for example NuScenes~\cite{caesar2020nuscenes}, neglecting nuanced human-LKA interactions and failing to represent diverse global driving conditions adequately. This shortcoming restricts researchers from conducting comparative studies between human-driven and automation-driven behaviors, essential for advancing safety and reliability. Additionally, most existing datasets are limited in environmental variability, insufficiently capturing adverse weather, degraded road infrastructure, and complex lighting conditions that are prevalent in real-world driving scenarios. Such limitations impede the development and validation of robust LKA systems capable of handling real-world operational complexities.

\section{Experiments, Data Collection and Processing}
\label{sec:experiments}

OpenLKA addresses the critical need for real-world, multimodal, scalable datasets for LKA research. Key design principles include \textbf{extensibility}, \textbf{openness}, \textbf{multimodal richness} and \textbf{contextual diversity}. Unlike existing datasets, OpenLKA integrates synchronized CAN bus, dashcam video, OP outputs, and contextual scene annotations, making it uniquely comprehensive.

\subsection{Experiment Hardware Setup}
\label{subsec:methods}

\begin{figure}[t]
  \centering
  \includegraphics[width=0.9\linewidth]{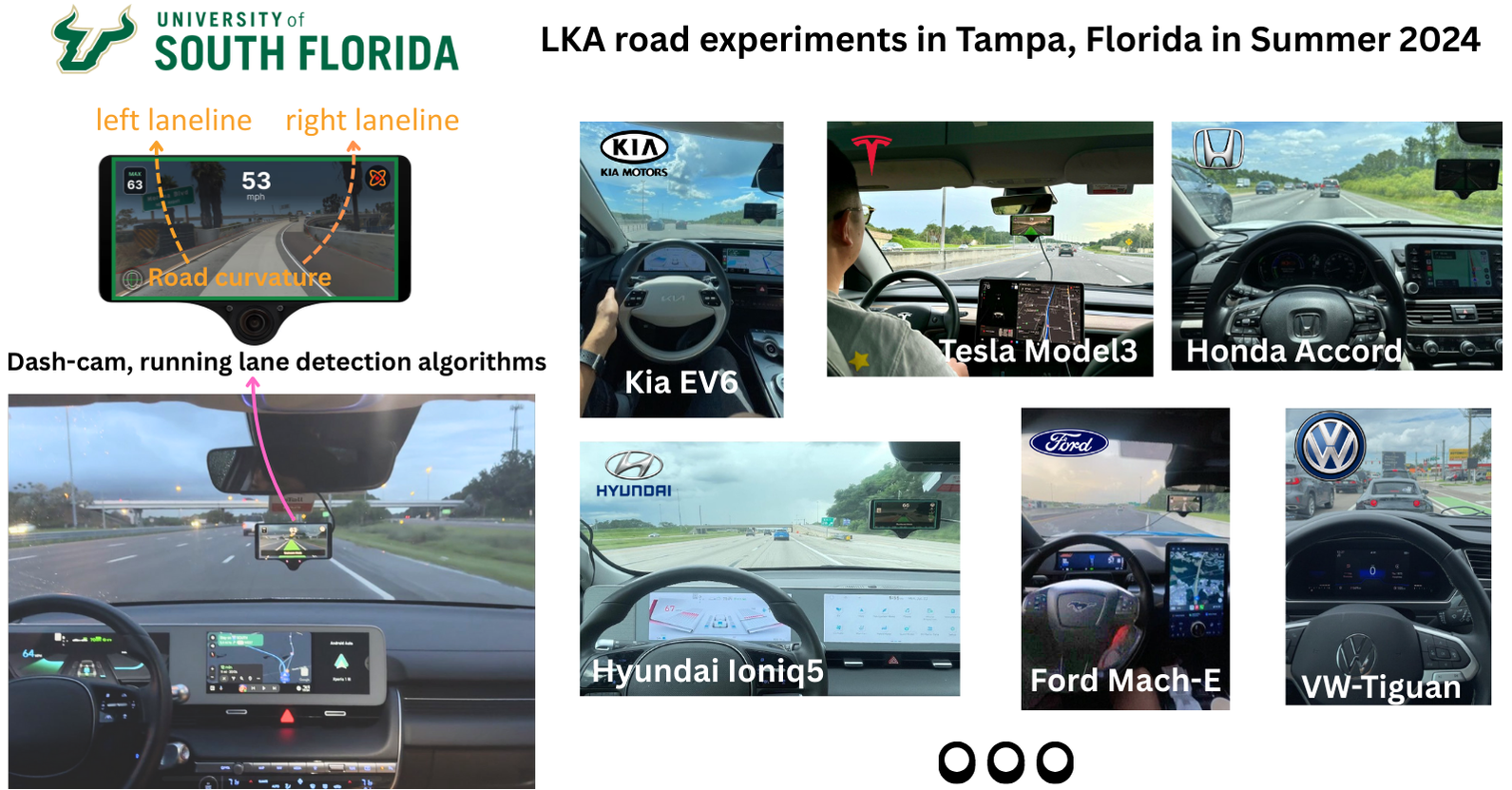}
  \caption{Experiment setup for LKA data collection and testing fleet using recent rental cars from several car makes.}
  \label{fig:comma_setup}
  \vspace{-0.5em}
\end{figure}

To evaluate the real-world performance of LKA systems in production vehicles~\cite{re2021testing, song2023identifying}, we utilized the Comma Three-X~\cite{comma2024}, an aftermarket self-driving development kit. The Comma Three-X was selected for its integrated high-definition front-view camera and access to the OP ecosystem, which facilitates synchronized data collection. The device connects to the vehicle's internal Controller Area Network (CAN) bus via the physical On-Board Diagnostics (OBD-II) port, enabling the extraction of high-frequency control signals through reverse engineering, as standard OBD-II protocols do not provide such detailed data.

The Comma Three-X was mounted in the center of the vehicle's interior windshield (Fig.~\ref{fig:comma_setup}) after rigorous calibration of its camera to ensure accurate lane deviation measurements. Its internal sensors operate at a frequency of 100 Hz, and its state-of-the-art lane detection algorithm, which employs an end-to-end approach based on computer vision and machine learning, identifies lane lines in real time. However, this algorithm serves as a reference and does not reflect the performance of the vehicle's native LKA system. By synchronizing CAN data with video footage, we obtained a comprehensive view of the vehicle's mechanical responses in relation to its driving environment.

To eliminate interference with the vehicle's native LKA system, the Comma Three-X was operated in \emph{dash-cam mode}, where it passively collects CAN bus data without exerting control. We confirmed the absence of interference through electrical load testing and monitoring of the vehicle's native system status, ensuring the integrity of the stock LKA system during data collection.

\subsection{Selection of Testing Vehicles and Scenarios}
\label{subsec:scenarios}

\begin{table*}[htbp]
\caption{Summary of Testing Scenarios}
\label{tab:testing_conditions}
\centering
\renewcommand{\arraystretch}{1.2}
\setlength{\tabcolsep}{6pt}
\small
\begin{tabular}{@{}l l@{}}
\toprule
\textbf{Variable} & \textbf{Conditions Tested} \\
\midrule
\textbf{Weather} & Clear, Rain, High Rain, Storm, Thunderstorms, Fog, Extreme Heat \\
\textbf{Lighting} & Daytime, Nighttime, Dawn, Dusk, Sun glare, Street lights, Tunnel transitions \\
\textbf{Pavement Quality} & Smooth asphalt, Rough surfaces, Potholes, Wet surfaces, Gravel, Repaired patches \\
\textbf{Lane Markings} & Well-defined, Faded, Obscured, Absent, Reflective markers, Different Color Combinations, Transitions \\
\textbf{Road Geometry} & Straight, Sharp curves, Grades, Intersections, Construction zones, Merges/splits, Roundabouts, Tunnels \\
\textbf{Traffic} & Free-flow, Light, Moderate, Heavy Congested, Stop-and-go, Traffic Jam \\
\textbf{Vehicle Speeds} & Low ($<$20 mph), Medium (20--45 mph), High ($>$45 mph), Variable speeds \\
\bottomrule
\end{tabular}
\end{table*}

Our experimental design aimed to comprehensively assess LKA performance under diverse real-world driving conditions. We conducted tests using a fleet of rental cars sourced from Tampa International Airport, featuring models from major manufacturers such as Honda, Toyota, Tesla, Ford, Kia, Hyundai and so on (Fig.~\ref{fig:comma_setup}). This selection ensured broad coverage of LKA implementations across different automakers.

\begin{figure}[htbp]
  \centering
  \includegraphics[width=0.46\textwidth]{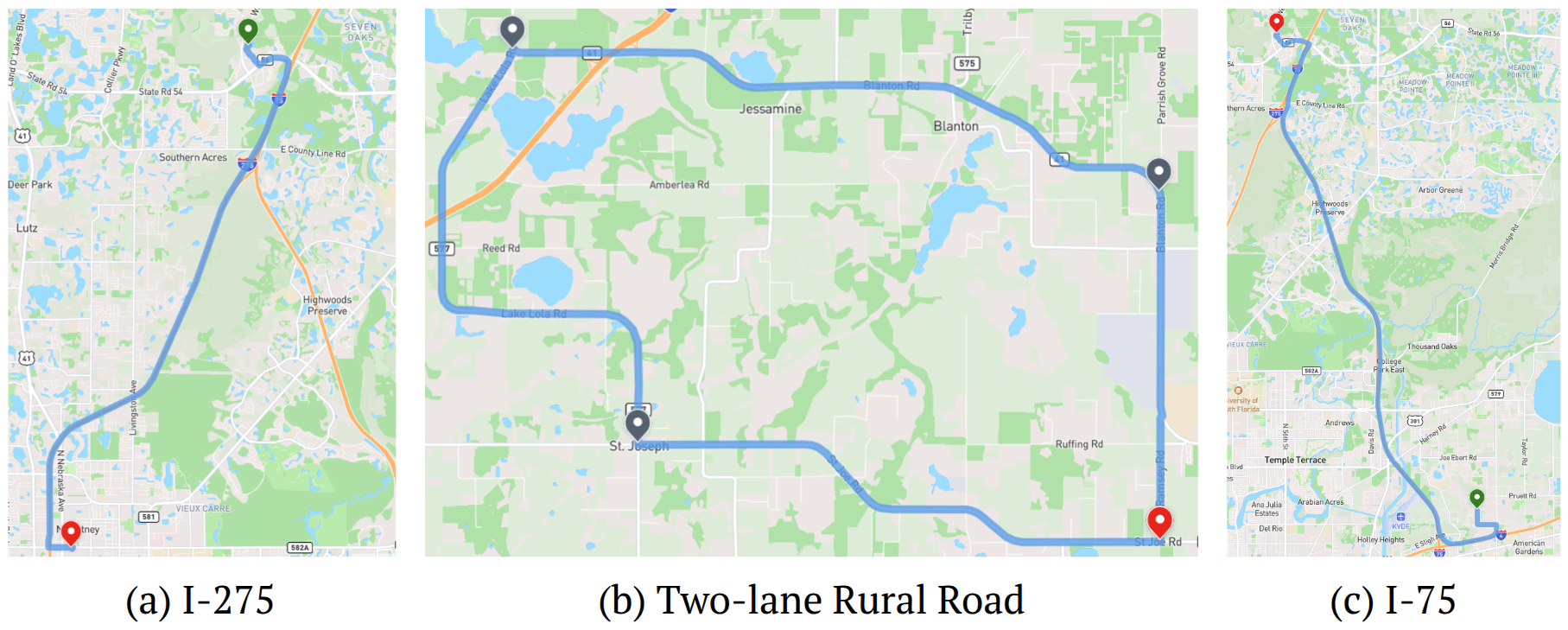}
  \caption{LKA data collection sites in Tampa, Florida, covering I-275, I-75, and two-lane, two-way rural road.}
  \label{fig:layouts}
  \vspace{-1em}
\end{figure}

\begin{figure}[htbp]
  \centering
  \includegraphics[width=0.46\textwidth]{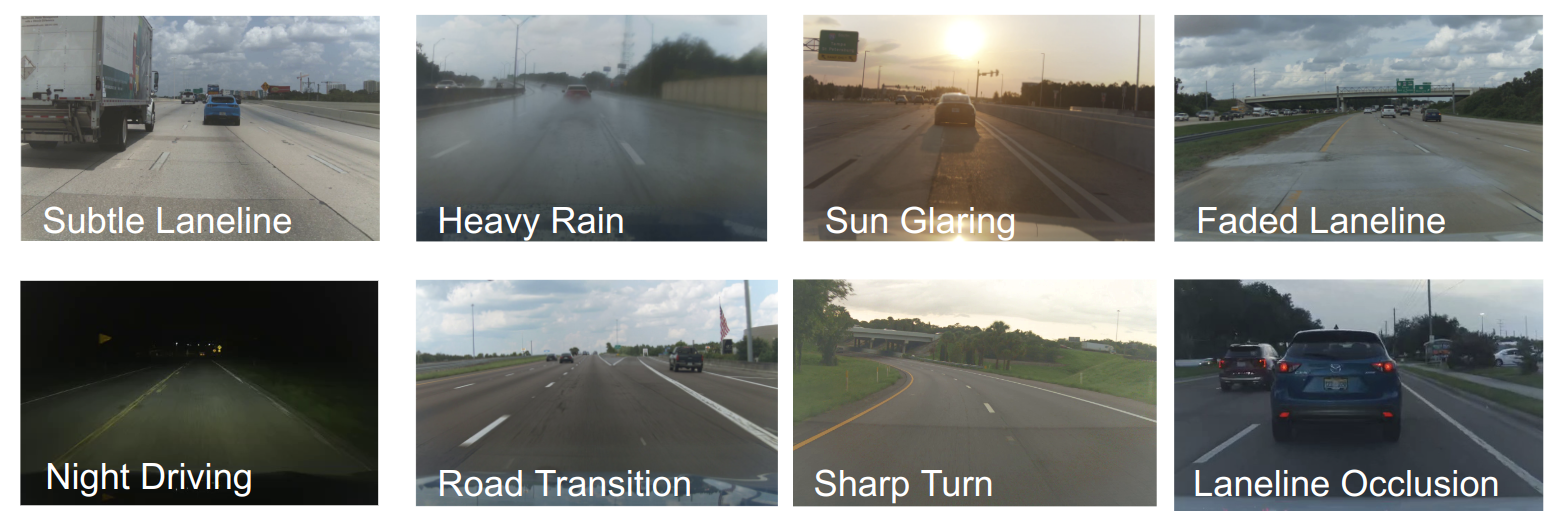}
  \caption{Examples of testing scenarios in OpenLKA. }
  \label{fig:diff_scenarios}
\end{figure}

Testing was primarily conducted in Tampa, Florida, and surrounding areas, leveraging the region's subtropical climate to evaluate LKA systems under challenging conditions, such as heavy rain, high humidity, sharp curves, unclear lane markings, and tunnels, summarized in Table~\ref{tab:testing_conditions}. Routes were systematically selected to include complex road segments (Fig.~\ref{fig:layouts}), and tests were performed across various times of day and weather conditions to ensure comprehensive coverage (Fig.~\ref{fig:diff_scenarios}). Additionally, we incorporated data from a community dataset, which, despite its randomized collection nature, captured diverse weather, traffic, and pavement conditions from multiple regions, further enhancing the dataset's diversity.

All tests were conducted by our team, consisting of experienced drivers with extensive driving backgrounds. We adhered to a protocol of utilizing the LKA system whenever feasible, provided it did not result in significant vehicle deviation or disengagement, ensuring consistent evaluation of the system's performance under varied conditions.


\subsection{CAN Log Data Acquisition and Reverse Engineering}
\label{subsec:can_processing}

\begin{figure}[htbp]
  \centering
  \includegraphics[width=0.48\textwidth]{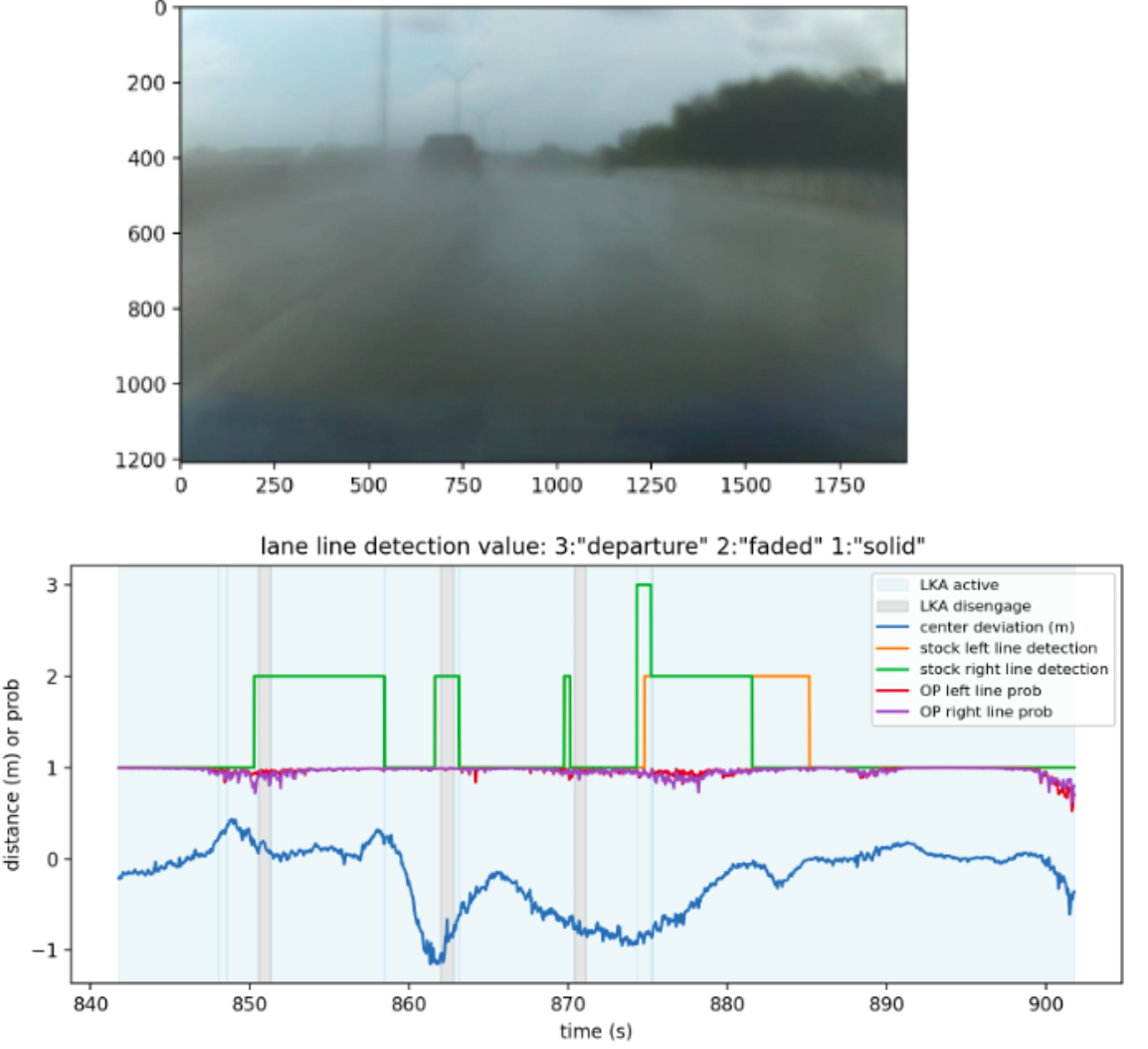}
    \caption{%
    CAN and video data sample collected in heavy rain:  
        (a) the front camera view.  
        (b) LKA-related CAN data in 60s, including lane deviation (blue), laneline detection result (green, orange; 1 solid, 2 faded, 3 departure), laneline detection confidences by OP (magenta, red), and LKA engagement: active (blue shading) vs.\ disengaged (grey).%
    }
  \label{fig:examples}
  \vspace{-1em}
\end{figure}

CAN logs were acquired using the Comma Three‑X by accessing the vehicle’s internal CAN bus through the OBD‑II port. The logs were recorded in two formats: qlog files (10 Hz, for low‑frequency state monitoring) and rlog files (100 Hz, for high‑frequency dynamic analysis). We initially employed open‑source DBC files from OpenDBC to decode CAN messages—extracting variables such as clock, LKA/ACC status, pose angles, steering metrics, lane detection, lane offset, object detection, dynamics, and jerk. The right part of Fig.~\ref{fig:examples} shows an example that we reversed engineer the laneline detection results and LKA status from Hyundai IONIQ5. The availability of signals per vehicle model is detailed in Table II. However, OpenDBC files only natively support a subset of vehicle makes (e.g., Ford, Toyota Camry, Corolla, Kia, Chevrolet). To extend support to remaining vehicle models, we performed additional work: (1) curated and integrated richer DBC definitions by retrieving files shared in public forums and inferring message structures from partially documented CAN signal information; (2) reverse‑engineered DBCs by adapting files from similar or earlier‑model years of the same make; and (3) applied a systematic process of coarse‑to‑fine message localization—grouping signals by message ID header, correlating with known vehicle behavior domains, and validating candidate signals via controlled vehicle observations. All resulting DBC files have been openly released in our GitHub repository to support future analysis and transparency. 

\begin{table}[htbp]
\caption{Signal Availability Across Vehicle Models}
\label{tab:signal_availability}
\centering
\footnotesize
\renewcommand{\arraystretch}{1.1}
\setlength{\tabcolsep}{2pt}
\begin{tabular}{@{}lcccccccccc@{}}
\toprule
\textbf{Model} & \textbf{Clk} & \textbf{ACC} & \textbf{Pose} & \textbf{Steer} & \textbf{Lane Det.} & \textbf{Obj. Det.} & \textbf{Dyn.} & \textbf{Jerk} \\
\midrule
Volkswagen & Y & Y & Y & Y & Y & Y & Y & Y \\
Ford & Y & Y & Y & Y & N & Y & Y & N \\
Toyota & Y & Y & Y & Y & Y & N & Y & N \\
Honda & Y & Y & Y & Y & Y & Y & Y & Y \\
Tesla & Y & Y & Y & Y (RE) & Y (RE) & Y & Y & Y \\
Kia & Y & N & Y & Y & N & Y & Y & N \\
\bottomrule
\end{tabular}
\caption*{\small Y: Yes, N: No, RE: Reverse Engineered.}
\end{table}

\subsection{Scalable and Robust Laneline Estimation via OP}
\label{subsec:op_integration}

We integrate OP, a state‑of‑the‑art, open‑source advanced driver‑assistance system, into our dash‑cam platform to record high‑fidelity perception outputs that are time‑aligned with the forward‑facing video.  OP’s vision model continuously estimates key lane‑keeping variables, including the vehicle’s lateral offset to each lane line and the instantaneous road curvature.

\begin{figure}[htbp]
  \centering
  \includegraphics[width=0.48\textwidth]{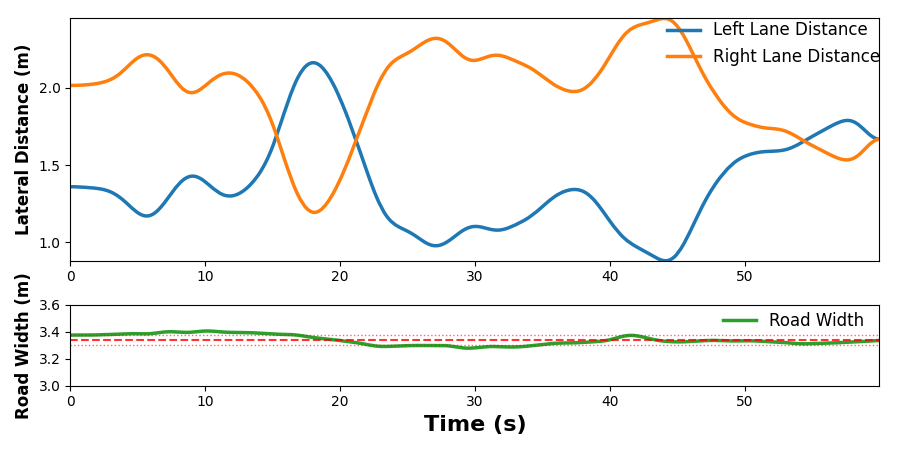}
  \caption{
    Top: OP‑estimated lateral distances from the vehicle centre to the left (blue) and right (orange) lane lines. 
    Bottom: Road width (green) obtained by summing the two distances; the dashed red line marks the nominal 3.35 m lane width. The width remains nearly constant, suggesting the robustness and accuracy of OP’s lane detection algorithms.%
  }
  \label{fig:OPverification}
  \vspace{-0.5em}
\end{figure}

Although not perfectly exact, these estimates provide a quite accurate, practical, scalable proxy for ground truth, avoiding the cost and labor of LiDAR‑based mapping or manual road‑geometry surveys.  OP thus enables efficient collection of supervisory signals across large‑scale, real‑world driving data.  Figure~\ref{fig:OPverification} plots the OP‑estimated distances to the left and right lane lines; their sum approximates the roadway width, which remains stable throughout the drive.  Thanks to OP’s end‑to‑end architecture, its lane‑detection confidence is consistently higher than that of the stock system, as shown in Fig.~\ref{fig:examples}.

Running OP in a passive, observational mode preserves the native behavior of the production LKA while simultaneously generating a consistent, model‑based perception stream.  This dual‑view setup enables side‑by‑side comparisons between proprietary controllers and a common reference model, supporting fine‑grained analyses of lane centering and lateral control.  The middle panels of Fig.~\ref{fig:examples} illustrate an example: OP directly yields the lateral distances from the vehicle center to the left and right lane boundaries (positive to the right), from which we compute the lane‑center deviation.  Manual inspection confirms that the temporal derivative of this deviation closely matches the measured lateral velocity, indicating strong internal consistency and high precision in OP’s perception outputs.  This empirical agreement justifies using OP estimates as a practical surrogate for ground‑truth lane position in large‑scale observational studies.

\subsection{VLM-Based Scene Annotations}
\label{subsec:video_augmentation}

\begin{table*}[htbp]
\centering
\begin{threeparttable}
\caption{Description of Annotated Labels}
\label{tab:label_description}
\footnotesize
\renewcommand{\arraystretch}{1.2}
\setlength{\tabcolsep}{5pt}
\begin{tabular}{@{}p{2.4cm}ccccccc@{}}  
\toprule
\textbf{Category} & \textbf{0} & \textbf{1} & \textbf{2} & \textbf{3} & \textbf{4} & \textbf{5} & \textbf{6} \\
\midrule
\textbf{Lane Mark.}  & Worn out & Partially worn & Intact & -- & -- & -- & -- \\
\textbf{Weather}     & Rain & Dusty & Clear & Snow & -- & -- & -- \\
\textbf{Lighting}    & Night & Night with light & Daylight & -- & -- & -- & -- \\
\textbf{Traffic}     & Traffic jam & Stop\textendash and\textendash go & Congested & Free flow & -- & -- & -- \\
\textbf{Road Cond.}  & Large potholes & Small cracks & Smooth & -- & -- & -- & -- \\
\textbf{Driving}     & Reckless & Aggressive & Normal & Timid & -- & -- & -- \\
\textbf{Pedestrian}  & None & Few & Moderate & Crowded & -- & -- & -- \\
\textbf{Road Type}   & Two\textendash Lane\tnote{a} & Three\textendash Lane\tnote{b} & Four\textendash Lane\tnote{c} & Multilane\tnote{d} & Laneline Transition & Intersection & -- \\
\textbf{Scenarios}   & None & Oncoming vehicle & Emergency vehicle & Construction & Obstacle & Accident & Others \\
\textbf{Surr. Veh.}  & None & Low density & Moderate density & High density & -- & -- & -- \\
\bottomrule
\end{tabular}
\begin{tablenotes}[flushleft]
\footnotesize
\item[a] \textbf{2U}: Two\textendash lane, two\textendash way undivided road.
\item[b] \textbf{3T}: Three\textendash lane, two\textendash way road with center two\textendash way left\textendash turn lane (TWLTL).
\item[c] \textbf{4U}: Four\textendash lane, two\textendash way undivided arterial.
\item[d] Multilane arterial with~$\ge{}5$ total lanes (e.g., 5U/6D).
\end{tablenotes}
\end{threeparttable}
\end{table*}


\begin{figure}[htbp]
  \centering
  \includegraphics[width=0.48\textwidth]{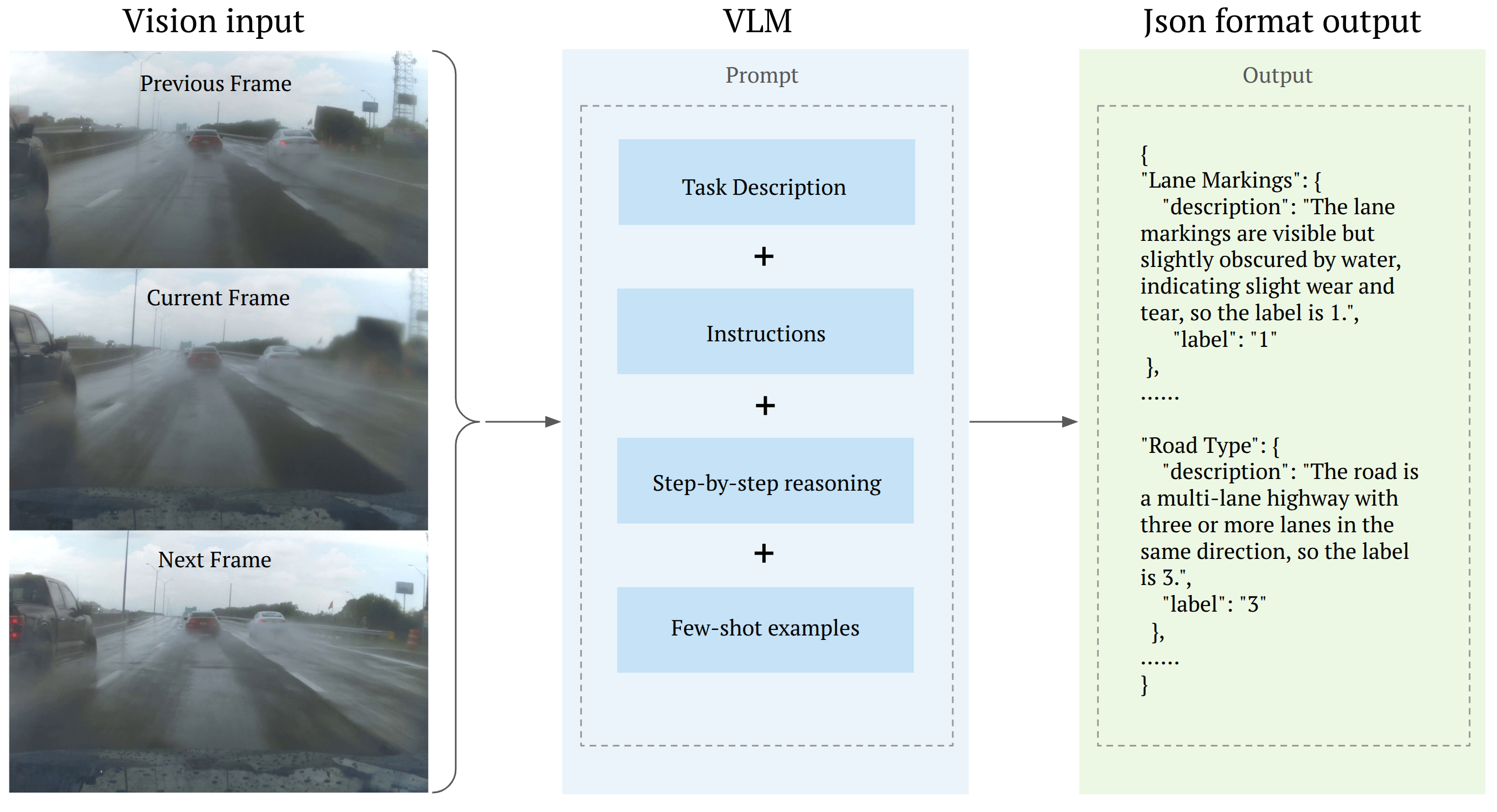}
  \caption{Annotation pipeline for video frames using VLM.}
  \label{fig:llm_annotation}
  \vspace{-0.5em}
\end{figure}

To provide richer contextual information for detailed analysis of LKA scenarios descriptions, we annotated the OpenLKA video data using VLM~\cite{tan2024large}, specifically GPT-4o. The video footage was segmented into individual frames, each analyzed together with its immediate preceding and subsequent frames. This method reduces visual obstructions (e.g., windshield wipers, water droplets) and enhances temporal context, leading to more robust and accurate annotations (see Fig.~\ref{fig:llm_annotation}).

\begin{figure}[htbp]
  \centering
  \includegraphics[width=0.48\textwidth]{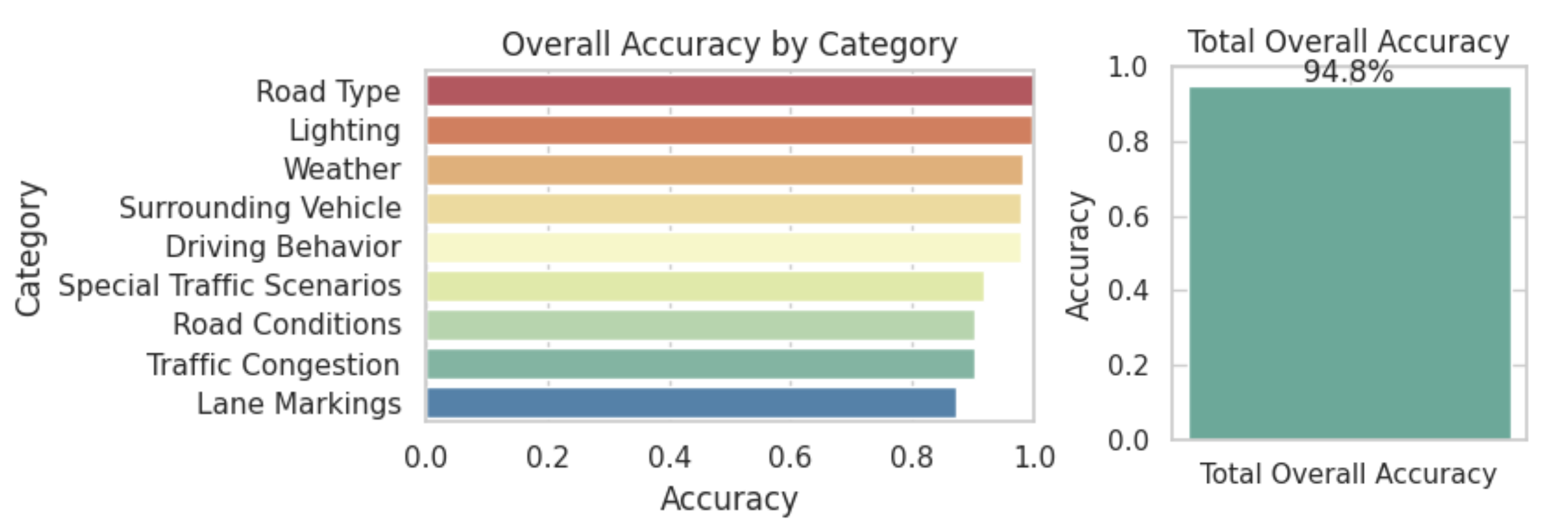}
  \caption{Comparision of human's label and VLM's annotation on sampled dataset.}
  \label{fig:label_compare}
  \vspace{-0.5em}
\end{figure}

We developed a domain-specific annotation prompt tailored explicitly for transportation scenarios within the OpenLKA dataset. The prompt utilized System Prompting, Role Prompting and few-shot prompting techniques, instructing the VLM to act as an experienced transportation domain expert~\cite{brown2020language}. Additionally, the prompt incorporated the Chain-of-Thought (CoT) methodology\cite{wei2022chain}, promoting systematic, step-by-step reasoning to improve consistency and reduce annotation hallucinations~\cite{wei2022chain}. Few-shot examples were also included to clearly illustrate the intended annotation process. The VLM classified each frame into predefined environmental categories (as shown in Table~\ref{tab:label_description}). Annotation accuracy was validated by manually labeling a representative subset of 2,000 images, selected via stratified random sampling across all categories. Comparing manual annotations with GPT-4o's outputs revealed an overall accuracy exceeding 94\% and each category's accuracy exceeding 85\%, shown in Fig.~\ref{fig:label_compare}. Notably, the VLM exhibited a conservative labeling tendency, frequently identifying scenarios as more severe than manual annotations, a beneficial characteristic for safety-critical transportation system design.

\subsection{Ecosystem with Community-Driven Data Streaming}

OpenLKA uniquely incorporates continuous updates from the global comma.ai self-driving community, marking it the largest LKA dataset to leverage community-sourced data streams. Our open-source pipeline automatically converts community-collected CAN/video data into OpenLKA-compatible formats. Users worldwide can seamlessly contribute data, and our processing tools (CAN decoding, synchronization scripts, VLM annotation methods) are fully open-sourced. We anticipate OpenLKA evolving into one of the most diverse LKA datasets globally, empowering researchers to create customized datasets using our publicly available tooling.

\section{Dataset Overview}

\begin{table*}[t]
\caption{Overview of OpenLKA Dataset Contents}
\label{tab:dataset_contents}
\centering
\small
\setlength{\tabcolsep}{4pt}
\begin{tabular}{@{}p{2.5cm} p{5cm} p{9cm}@{}}
\toprule
\textbf{Category} & \textbf{Component} & \textbf{Description} \\
\midrule
\multirow{3}{*}{Video Data} 
& Low-Resolution Video  & Standard front view, 526$\times$330 resolution, available for all routes. \\
& High-Resolution Video  & Detailed front view, 1928$\times$1208 resolution, for self-collected subset. \\
& Fisheye Video  & Wide-angle view, 1928$\times$1208 resolution, for self-collected subset. \\
\midrule
\multirow{2}{*}{Decoded CAN} 
& LKA Status & LKA System engage status, Lane depature warning output, OEM detection outputs \\
& Vehicle Dynamics & Motion data including speed, acceleration, steering, and body orientation. \\
\midrule
\multirow{2}{*}{\parbox{2.5cm}{Vision Detection Model(OP) Outputs}}
& Perception Outputs & Environmental awareness data, e.g. lane visibility, lane departure risks, and lead vehicle detection. \\
& Planning Outputs & Path planning, e.g. desired curvature and predicted trajectories. \\
\midrule
VLM Annotations & annotation.json & Environmental factors: weather, traffic congestion, lighting, road markings. \\
\midrule
\multirow{2}{*}{Other} 
& DBC Files & Reverse-engineered CAN DBC files for various vehicle models. \\
& Prompt Template  & Chain-of-Thought prompt for VLM annotation. \\
\bottomrule
\end{tabular}
\end{table*}

\begin{figure}[htbp]
  \centering
  \includegraphics[width=0.46\textwidth]{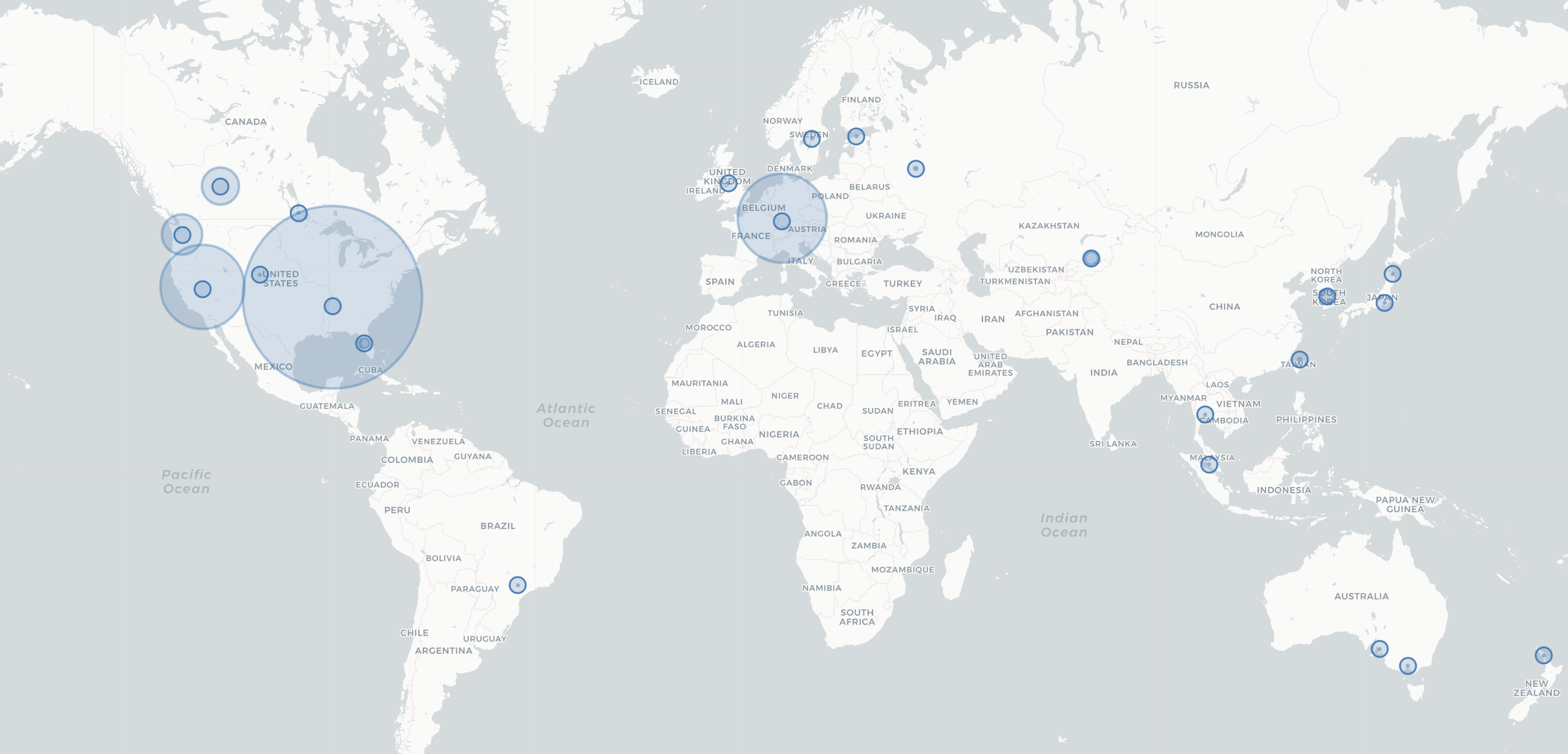}
  \caption{World-wide open driving logs from community}
  \label{fig:HVdataset}
  \vspace{-1em}
\end{figure}

The OpenLKA dataset was developed to address the need for a comprehensive, multimodal resource tailored for LKA system research, offering a diverse and parameter-rich foundation for analyzing real-world LKA performance and human-LKA interactions. Spanning 389.1 hours across 1,166 unique driving routes, the dataset encompasses data from 62 vehicle models and 52 drivers, organized in a structured \texttt{car model, dongle id, route id} format to distinguish between models, drivers, and trips. In the first, self-collected LKA data, was gathered starting in May 2024 by three drivers using 19 market-leading vehicle models rented from Tampa International Airport, covering a wide range of brands prevalent in the U.S. market, resulting in 323 hours (83\% of the total duration) of driving data across Florida’s diverse weather, lighting, road, and traffic conditions; the second component, sourced from the Comma autonomous driving community, comprises 66.1 hours (17\%) of human driving data from 49 drivers across 50+ vehicle models worldwide, collected as of April 2025, providing a valuable baseline for comparing human driving styles with LKA performance and exploring human-LKA interaction dynamics, as illustrated by the global distribution in Figure~\ref{fig:HVdataset} \cite{comma2024}. Additionally, this part of the dataset covers more than 30 states, provinces, and territories in over 80 cities around the world. This greatly enriches the dataset with driving scenarios such as snow, dust storms, and frosty roads, as well as driving scenarios in mountainous areas, dense cities, and deserts.

The dataset’s diversity is evident in its composition, the most represented model, HYUNDAI IONIQ 5 2022, contributes 14.61\% of the total duration (56.85 hours). The dataset includes synchronized multimodal streams, such as videos at multiple resolutions, vehicle control data via CAN, OP’s end-to-end model outputs, and VLM-annotated environmental factors, with a detailed breakdown provided in Table~\ref{tab:dataset_contents}. Privacy is ensured by blurring faces and license plates in all videos to protect driver identities.

Table~\ref{tab:comparison} compares OpenLKA with existing autonomous driving datasets, underscoring its unique strengths: comprehensive environmental annotations covering weather, visibility, traffic, and infrastructure; synchronized video and CAN data for precise temporal alignment; rich CAN signals paired with OP’s baseline outputs; and VLM-augmented labels that enhance contextual understanding, positioning OpenLKA as a uniquely diverse and parameter-rich resource for advancing LKA system development and dataset-driven research in autonomous driving.

\begin{table}[htbp]
\caption{Comparison of OpenLKA and Existing Datasets}
\label{tab:comparison}
\centering
\small
\setlength{\tabcolsep}{3pt}
\begin{tabular}{@{}p{2cm} ccc cccc c c@{}}
\toprule
\textbf{Dataset} & \multicolumn{3}{c}{\textbf{Data}} & \multicolumn{4}{c}{\textbf{Env}} & \textbf{LLM} & \textbf{Time} \\
\cmidrule(lr){2-4} \cmidrule(lr){5-8} \cmidrule(lr){9-9} \cmidrule(lr){10-10}
 & Vis & CAN & LiD & Wthr & Vis & Traf & Infra & Ann & (hrs) \\
\midrule
\textbf{OpenLKA} & \textbf{1} & \textbf{1} & 0 & \textbf{1} & \textbf{1} & \textbf{1} & \textbf{1} & \textbf{1} & \textbf{389.1} \\
HDD & 1 & 0 & 0 & 1 & 1 & 1 & 1 & 0 & 104 \\
BDD-X & 1 & 0 & 0 & 1 & 1 & 1 & 1 & 1 & 77 \\
WoodScape & 1 & 0 & 0 & 1 & 1 & 1 & 1 & 0 & N/A \\
DAWN & 1 & 0 & 0 & 1 & 1 & 1 & 1 & 0 & N/A \\
KITTI & 1 & 0 & 1 & 0* & 0 & 0 & 0 & 0 & 1.5 \\
BDD100K & 1 & 0 & 0 & 1 & 1 & 1 & 1 & 0 & 1.2K \\
Comma2k19 & 1 & 1 & 0 & 0 & 0 & 0 & 0 & 0 & 33 \\
Cityscapes & 1 & 0 & 0 & 1 & 1 & 1 & 1 & 0 & N/A \\
ApolloScape & 1 & 0 & 1 & 1 & 1 & 1 & 1 & 0 & N/A \\
nuScenes & 1 & 0 & 1 & 1 & 1 & 1 & 1 & 0 & 1.2K+ \\
Waymo & 1 & 0 & 1 & 1 & 1 & 1 & 1 & 0 & 574 \\
\bottomrule
\end{tabular}
\caption*{\small *Limited weather conditions in KITTI dataset.}
\end{table}

By uniting production‑vehicle data collection with tightly synchronized video, CAN, OP outputs, and VLM scene labels, OpenLKA delivers a cost‑efficient yet richly annotated mirror of everyday lane‑keeping.  This breadth of signals and operating conditions not only establishes a realistic benchmark for today’s LKA systems, but also unlocks a wide spectrum of downstream studies—from algorithm refinement to infrastructure evaluation and human–machine interaction analysis.  The following Discussion illustrates these opportunities and charts concrete research directions enabled by the dataset.

\section{Discussion: Dataset Applications}
\label{sec:applications}

The OpenLKA dataset, with its rich multimodal and diverse real-world driving data, offers a robust foundation for advancing research in intelligent transportation systems (ITS) and autonomous driving (AD). Below, we outline several potential applications that leverage the dataset's unique characteristics.

\begin{figure}[htbp]
  \centering
  \includegraphics[width=\linewidth]{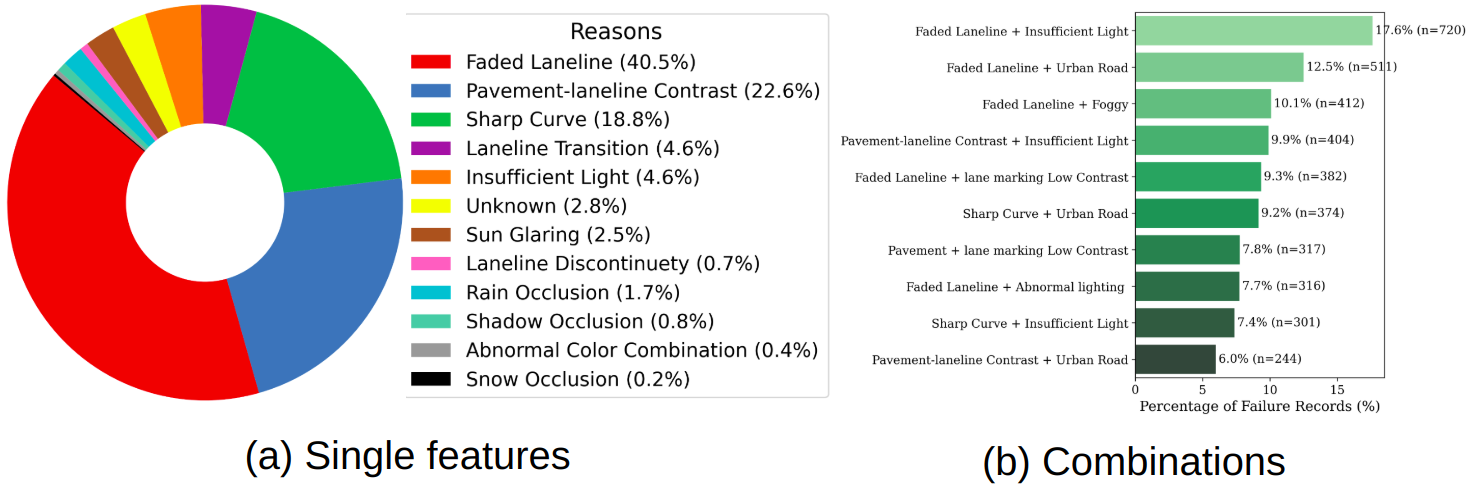} 
  \caption{Dominant causes of LKA failure extracted from OpenLKA:
  (a) share of single‑factor failures; (b) top multi‑factor combinations and their prevalence.}
  \label{fig:failure_modes}
  \vspace{-1em}
\end{figure}

\begin{figure}[htbp]
  \centering
  \includegraphics[width=0.48\textwidth]{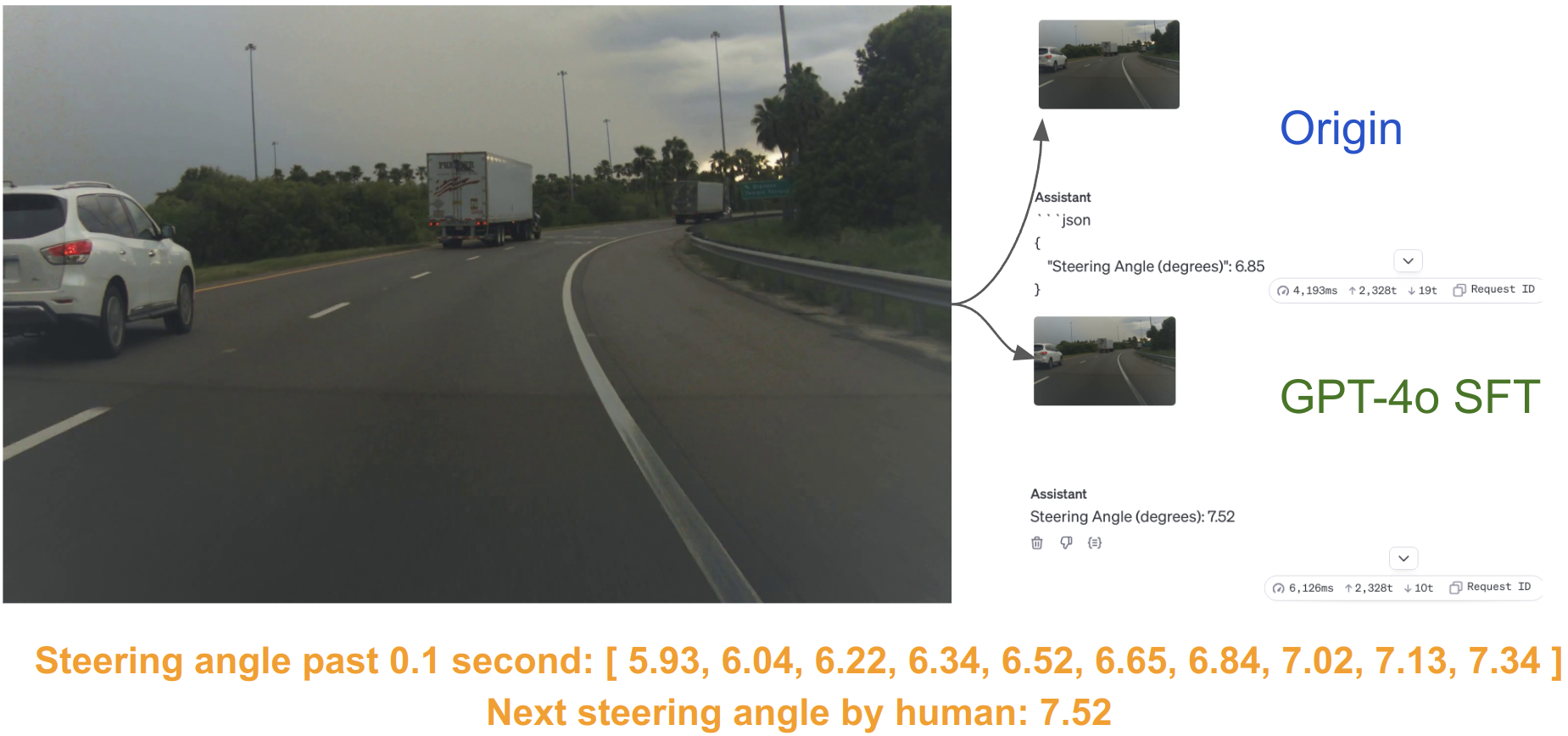}
  \caption{Steering angle prediction using GPT-4o: The original model predicts 6.85\textdegree, while fine-tuning with OpenLKA data yields 7.52\textdegree, matching the human driver’s actual action.}
  \label{fig:steering_angle}
  \vspace{-0.5em}
\end{figure}

\begin{figure}[htbp]
  \centering
  \includegraphics[width=0.3\textwidth]{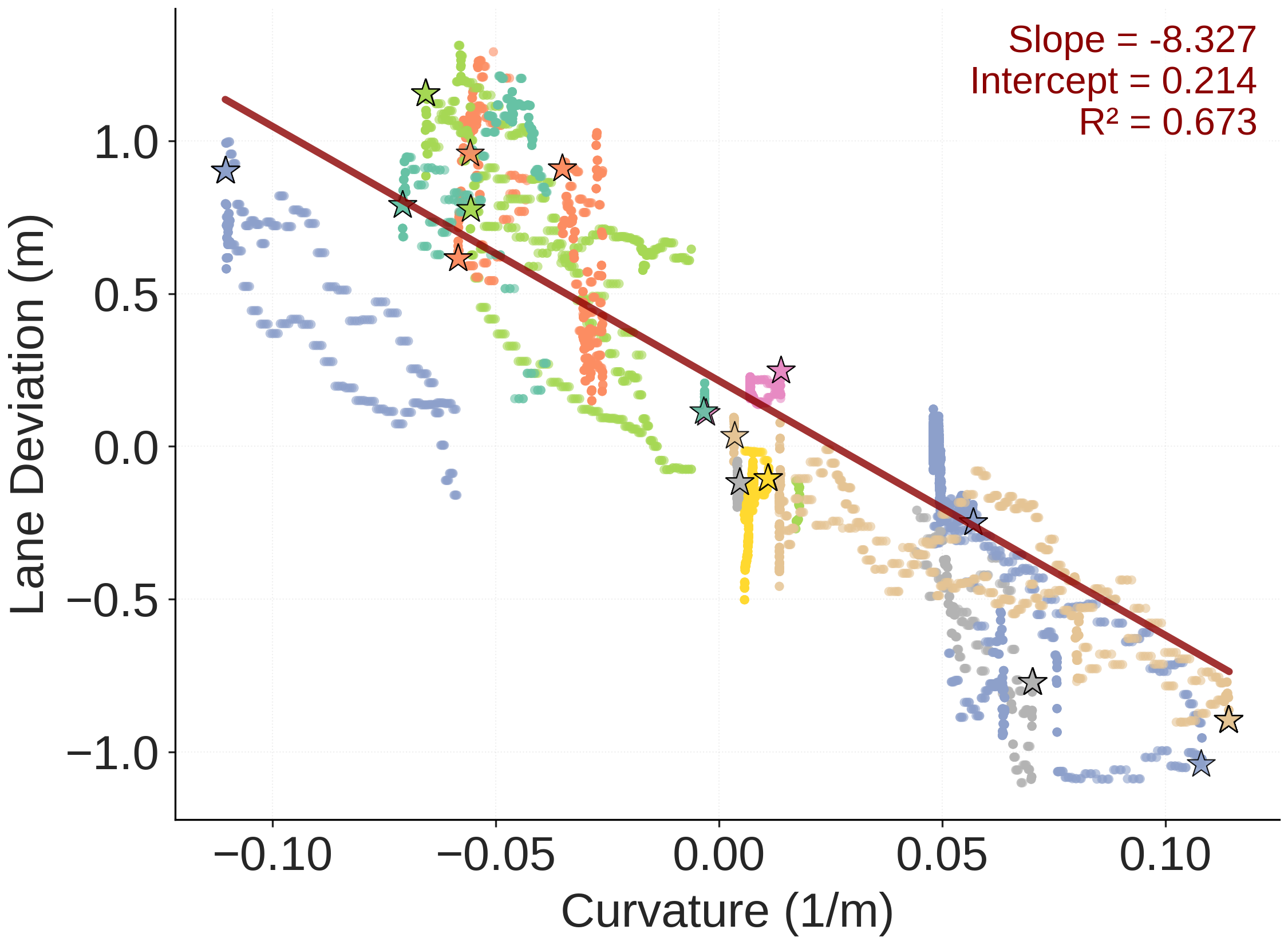}
  \caption{Quantitative analysis of LKA's lane deviation and road curvature; the lane deviation increases with increasing road curvature, revealing an approximately linear relationship.}
  \label{fig:curvature}
  \vspace{-0.5em}
\end{figure}

\begin{enumerate}

\item \textbf{LKA benchmarking and real-world evaluation:}  

This real‑world dataset spans multiple vehicle brands and operating conditions, allowing us to (i) benchmark Lane‑Keeping Assist (LKA) performance by make and (ii) isolate the scenarios most prone to failure. Figure.\ref{fig:failure_modes} ranks the leading single factors—such as faded lane markings, poor pavement‑lane contrast, and sharp curvature—and their high‑impact combinations, revealing that visibility‑related degradations account for more than 60\% of recorded disengagements. These findings offer concrete guidance for both OEM algorithm refinement and infrastructure interventions, which we will elaborate in forthcoming work.

\item \textbf{Multimodal Foundation Models \& VLA:}  
Leveraging frame‑aligned video, CAN dynamics, and OP trajectories, \textit{OpenLKA} is a turnkey test bed for Vision–Language–Action (VLA) research\cite{jiang2025survey}.  Our preliminary experiments with multimodal foundation models (Fig.~\ref{fig:steering_angle}) show promising gains in steering prediction, and we will continue exploring this avenue for next‑generation LKA and VLA policy design.

\item \textbf{Infrastructure Readiness Assessment and Roadway Safety Planning:}  
OpenLKA lets infrastructure stakeholders diagnose where and why LKA systems falter. 
Our current work already quantifies how roadway factors—e.g., the relationship between lane deviation and curvature (in Fig.~\ref{fig:curvature}) shape LKA performance, and fuller results will appear in a follow‑up paper. Building on these analyses, practitioners can pinpoint segments with frequent disengagements, then direct counter‑measures such as pavement re‑striping, lane‑marking renewal, or geometric redesign to boost assisted‑driving safety.

\item \textbf{Analysis of LKA Failure to Driver Takeover Dynamics:}  
When LKA systems fail or exhibit significant lane deviation, drivers often have little to no time to react, posing safety risks. OpenLKA provides a comprehensive record of the entire process from LKA failure or large lane deviation to driver takeover, capturing multimodal signals. This enables researchers to analyze LKA failure modes, quantify the time window from failure to takeover, and model driver reaction patterns under diverse conditions~\cite{schwager2024analysis}. Such insights can inform the development of proactive warning systems that predict LKA instability and alert drivers in advance, enhancing human-machine interaction safety.

\end{enumerate}

\section*{VI. Conclusion}
\label{sec:conclusion}


The OpenLKA dataset pioneers a uniquely comprehensive, multimodal, and real-world foundation for advancing LKA system research. Its primary strength lies in its ground truth of decoded CAN signals that reveal real-world LKA performance, supplemented by synchronized dashcam videos, and extensive contextual annotations via VLM. Furthermore, OpenLKA uniquely leverages continuous data contributions from a large, global driving community, ensuring ongoing dataset expansion and diversity. 

The dataset still lacks heavy-duty vehicles, whose steering challenges and lane departure consequences can be even more critical than passenger cars. Future work on this dataset will focus on extension to commercial trucks. 



\bibliographystyle{IEEEtran}
\bibliography{references}
\end{document}